\documentclass[journal,twoside,web]{ieeecolor}

\usepackage{mathtools}
\usepackage{amsmath}
\usepackage{amsfonts}
\usepackage{amssymb}
\usepackage{dsfont}
\usepackage{graphicx}
\usepackage{caption}
\usepackage{subcaption}
\usepackage{float}
\usepackage{wrapfig}
\usepackage{tabularx}
\usepackage{booktabs}
\usepackage{multirow}
\usepackage{array}
\usepackage{ragged2e}
\usepackage{algorithm}
\usepackage{algorithmic}
\usepackage{xcolor}
\usepackage{soul}
\usepackage{ulem}
\usepackage{lipsum}
\usepackage{hyperref}
\usepackage{verbatim}

\usepackage{lineno}
\usepackage{url}
\usepackage{cite}
\usepackage{jsen}

\hypersetup{
colorlinks=true,
bookmarksopen=true,
bookmarksnumbered=true,
citecolor=red,
urlcolor=red,
}

\newcolumntype{Y}{>{\centering\arraybackslash}X}
\newcolumntype{L}{>{\RaggedRight\arraybackslash}X}
\newcolumntype{R}{>{\RaggedLeft\arraybackslash}X}

\def\BibTeX{{\rm B\kern-.05em{\sc i\kern-.025em b}\kern-.08em
T\kern-.1667em\lower.7ex\hbox{E}\kern-.125emX}}

\definecolor{abstractbg}{rgb}{0.89804,0.94510,0.83137}
\title{Reliable Piezoresistive Strain Sensing Through Physical Limits and Uncertainty Monitoring

}

\author{
Carmen Ballester,
Víctor Muñoz,
Dorin Copaci,
and Dolores Blanco
\thanks{Departamento de Ingeniería de Sistemas y Automática,}
\thanks{Universidad Carlos III de Madrid, Leganés, Spain.}
\thanks{E-mail: cballest@ing.uc3m.es}\thanks{This work has been submitted to the IEEE for possible publication. Copyright may be transferred without notice, after which this version may no longer be accessible.}}

\begin{document}
\thispagestyle{empty}

\IEEEtitleabstractindextext{
\fcolorbox{abstractbg}{abstractbg}{
\begin{minipage}{\textwidth}
\begin{wrapfigure}[20]{r}{3in}

\includegraphics[width=3in]{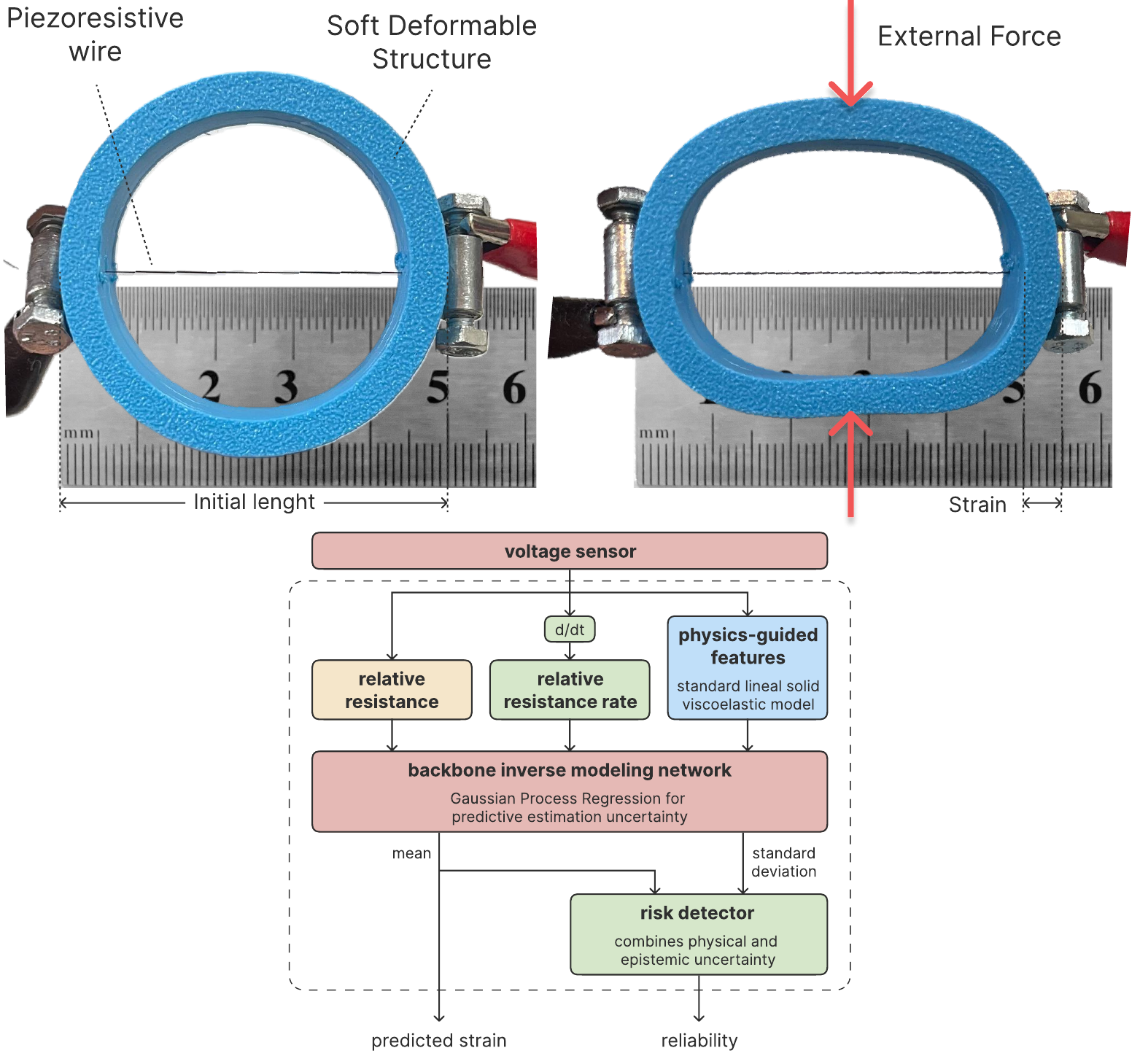}

\end{wrapfigure}

\begin{abstract}
Soft piezoresistive strain sensors are one of the most common sensing solutions for wearable and soft robotic applications due to their flexibility and compliance.
However, their resistance response is nonlinear and hysteretic, and a sensor can be pushed past its calibrated workspace or misbehave inside it, carrying that error into a decision or control loop.
Probabilistic regressors track confidence but ignore those limits. A predictive mean can look unremarkable even when the reading comes from a sensor outside its admissible range or already failing internally, so a confident-looking estimate is not the same as a trustworthy one.
This paper proposes a reliability framework pairing a physics-informed probabilistic inverse model, built on physics-guided input features, with a risk factor fusing uncertainty with strain and strain-rate limits into a three-state monitor. Tests on a Nitinol wire and a silver-coated polyamide thread with a Gaussian Process raised fit scores to 0.90-0.95 (RMSE 0.26\%-0.15\%) and a 96\% empirical coverage against the 95\% target. The monitor caught 95\% of out-of-range and 100\% of abnormal conditions while staying reliable under nominal operation.
A sensor that reports confidence alongside its estimate lets a system withhold action instead, since it needs no labeled failure examples, which are hard to collect for soft materials.
\end{abstract}


\end{minipage}}}

\maketitle

\section{Introduction}

\IEEEPARstart{S}{mart} materials are increasingly being adopted in several engineering applications due to their ability to directly respond to mechanical, electrical, thermal, or magnetic stimuli \cite{yadong2022}. In particular, piezoresistive sensors have emerged as one of the most widely adopted sensing solutions for soft robotic systems due to their simple readout electronics, low cost, lightweight nature, and ease of integration into deformable structures \cite{saxena2023}. Unlike conventional rigid sensors, piezoresistive materials can be directly embedded within soft bodies, enabling distributed sensing while preserving the mechanical compliance of the system \cite{qu2024}. Several materials have been explored for piezoresistive soft sensing. Conductive textiles and yarns are among the most common solutions for wearable applications because they can be naturally integrated into garments and conform to the human body \cite{choudhry2020}. Metallic fibers, including superelastic Nitinol wires, have also attracted attention due to their combined sensing and mechanical capabilities, allowing strain to be inferred from changes in electrical resistance \cite{srivastava2016}. In addition, conductive polymers and nanocomposite materials have become increasingly popular because they can simultaneously provide high sensitivity, large deformation ranges, and straightforward fabrication processes \cite{tazwar2025}. 

Despite these advantages, the behavior of piezoresistive soft sensors remains challenging to model. Their electrical response is affected not only by the current deformation state but also by hysteresis, viscoelasticity, strain-rate effects, drift, fatigue, and material degradation \cite{mostaghniyazdi2025}. These challenges become even more pronounced when the sensors are subjected to varying operating conditions, repeated loading cycles, or environmental disturbances.

Machine learning techniques have become increasingly popular for modeling soft sensors. Data-driven methods are capable of learning complex nonlinear mappings directly from experimental data and have demonstrated excellent performance in estimating physical quantities such as strain, curvature, force, and shape \cite{chen2025}. Neural Networks \cite{salayman2025}, Support Vector Machines \cite{tengfei2026}, Gaussian Processes \cite{musa2026}, and other nonlinear regression techniques such as transformers \cite{munoz2025} have all been successfully applied to soft sensing modeling problems.

The main limitation of the data-driven methods is that prediction accuracy alone does not guarantee measurement reliability \cite{costa2023}. Most works evaluate performance using metrics such as Root Mean Squared Error (RMSE) \cite{saha2023}, Mean Absolute Error (MAE) \cite{torkaman2023}, correlation coefficients \cite{alvarez2022}, or goodness-of-fit scores \cite{martindale2026}. While these metrics quantify the average prediction error on a test dataset, they provide little information about whether the resulting measurement should be trusted during deployment. 

One common approach consists of collecting examples of faulty behavior and training dedicated fault-detection algorithms \cite{yu2022}, \cite{cen2022}. These methods can successfully identify predefined failure scenarios when sufficient labeled data are available \cite{trapani2023}, or when sensor faults can be artificially injected into the measurement data \cite{li2023}. However, obtaining representative fault datasets is often expensive or impractical \cite{jandaghi2023}. Moreover, simulating failure scenarios often requires reproducing rare operating conditions that may not be easily observed during normal operation \cite{loo2021}.

An alternative to supervised fault detection algorithms are uncertainty-aware machine learning models \cite{ding2021}. Probabilistic models, such as Gaussian Processes \cite{jing2025} and Bayesian neural networks \cite{sahin2026}, estimate the output and the epistemic uncertainty associated with that prediction. In principle, this uncertainty can provide valuable information about whether the model is operating within regions that are well represented by the training data or if it is working outside of domain \cite{lin2025}.

When applying this approach to sensors, the unconsidered problem is that epistemic uncertainty alone does not necessarily capture whether the predicted sensor state remains physically plausible with the sensor's operating limits.

The reliability problem becomes especially critical when the sensor operates outside the conditions represented in the training dataset. In these situations, machine-learning models may be forced to extrapolate beyond their previous experience, producing predictions whose accuracy is unknown. Unexpected disturbances, changes in operating conditions, or hardware failures can all generate measurements that differ substantially from the data used during training. This is particularly relevant in closed-loop control applications, where incorrect sensor measurements may directly affect the stability, safety, and performance of the overall system.

In this work, we propose a reliability monitoring framework for piezoresistive soft sensors that combines physics-guided features, probabilistic inverse modeling, and reliability monitoring. Specifically, a physics-informed Gaussian Process turns strain estimation into a self-monitoring process, where at every measurement, it reports a strain value together with a reliability state. The complete framework is validated on two piezoresistive materials with different transduction mechanisms, a superelastic Nitinol wire and a silver-coated polyamide thread, each tested under nominal, out-of-range, and abnormal operating conditions.

Unlike fault-classification approaches, the proposed method does not require labeled failure data or examples of damaged sensors. Instead, reliability is inferred from the consistency between the model uncertainty and the sensor's admissible physical operating domain.

\section{Sensor's Physical Domain}\label{h1:physical-domain}

The characterization of a sensor is performed by subjecting it to a series of loading and unloading cycles with a maximum strain amplitude \( A_i \) and frequency \( f_j \). The set of all possible conditions \( C_{ij}(A_i, f_j) \) up to sensor failure defines its physical domain. The operational domain is a subset of this physical domain, since the sensor must satisfy the specified physical and quality constraints.

Physical constraints are evaluated at each timestep and impose hard limits on the sensor state. Each physical constraint is a function \( g_k \) of the current strain \( \epsilon(t) \) and strain rate \( \dot{\epsilon}(t) \) that must remain below a threshold \( \theta_k \), as in Equation \eqref{eq:physical-constraint}.

\begin{equation}
\begin{aligned}
\label{eq:physical-constraint}
g_k \left( \epsilon(t), \dot{\epsilon}(t) \right) \le \theta_k, \quad k = 1, \ldots, K
\end{aligned}
\end{equation}

Quality constraints are evaluated at the condition level and assess how well the sensor's resistance response follows a specific score metric, where the score \( G_m \) must exceed a threshold \( \theta_m \), as defined in Equation \eqref{eq:quality-constraint}.

\begin{equation}
\begin{aligned}
\label{eq:quality-constraint}
G_m \left( C_{ij} \right) \ge \theta_m, \quad m = 1, \ldots, M
\end{aligned}
\end{equation}

The constraints therefore partition the sensor's physical domain into four regions based on whether each constraint is satisfied:

\begin{itemize}
\item[-] Physical and quality constraints met: nominal region, used for training and operation.
\item[-] Physical constraints met, quality constraint not met: degraded region, where the sensor's response fails the quality score threshold.
\item[-] Physical constraints not met, quality constraint met: out-of-range region, where the sensor exceeds its physical operating limits.
\item[-] Neither constraint met: rejected region, where both constraint types are violated.
\end{itemize}

This partitioning divides the condition into a nominal region and non-nominal behavior. Since both constraint types are expressed in the physical domain, the machine learning input must be designed so that its structure reflects the physical domain boundaries.

\section{Uncertainty-Aware Sensor Modeling}

The relative resistance, \( \tilde{R}(t) \), depends on both the current strain and the sensor's recent loading history. The feature vector must capture both the sensor's dynamics and its instantaneous value. The regressor must also provide a strain estimate together with a calibrated measure of its predictive uncertainty. This uncertainty only reflects how closely the current input matches the training data, not whether the sensor's underlying physical state is admissible. Reliable sensor operation therefore requires combining the model's predictive uncertainty with an independent assessment of physical plausibility.

\subsection{Physics-Guided Input Features}\label{h2:features}

The piezoresistive sensor is modelled as a viscoelastic process because it exhibits both an instantaneous electrical response and relaxation with short-term memory effects. The Standard Linear Solid (SLS) model is adopted because it is the simplest viscoelastic model that captures both the instantaneous response and the time-dependent relaxation \cite{elsahy2023}. In the physical SLS analogy, one elastic branch acts immediately while a second branch relaxes over time. In this work, this structure is used only to define a causal memory state \( z(t) \), in \eqref{eq:sls-state}, that follows the measured resistance with time constant \( \tau \).

\begin{equation}
\begin{aligned}
\label{eq:sls-state}
\dot{z}(t) = \frac{1}{\tau}(\tilde{R}(t) - z(t))
\end{aligned}
\end{equation}

The SLS model is used to derive the memory feature \( z(t) \) that carries the viscoelastic history of the sensor. Without it, two timesteps with the same \( \tilde{R} \) but different loading histories would map to identical inputs despite producing different strains. For real-time implementation, the memory state is updated with a sample period \( \Delta t \), as in \eqref{eq:sls-state-discrete}, where \( \phi \) controls how quickly the memory state follows the measured resistance.

\begin{equation}
\begin{aligned}
\label{eq:sls-state-discrete}
z_{k+1} = \phi z_k + (1-\phi) \tilde{R}_{k}, \quad \phi = e^{-\Delta t / \tau}
\end{aligned}
\end{equation}

The input feature vector, shown in Equation \eqref{eq:input-vector}, guides the model with static information with \( \tilde{R}(t) \), dynamic information with the relative resistance rate \( \dot{\tilde{R}}(t) \), and memory with \( z(t) \).

\begin{equation}
\begin{aligned}
\label{eq:input-vector}
x(t) = \begin{bmatrix}\tilde{R}(t) & \dot{\tilde{R}}(t) & z(t)\end{bmatrix}^T \in \mathbb{R}^3
\end{aligned}
\end{equation}

\subsection{Gaussian Process Regression}\label{h2:gp}

The backbone network for the inverse modeling regression is a Gaussian Process (GP), because it predicts both strain and epistemic uncertainty, which depends on how well the current sensor state is represented in the training data. When the input \( x(t) \) lies in regions densely covered by nominal observations, the posterior is well-constrained and the uncertainty remains low. When the sensor enters poorly represented regions, such as under degradation or abnormal operating conditions, the posterior uncertainty increases, directly relating the feature domain and the physical domain.

The GP learns the resistance-to-strain mapping from the physics-guided feature vector \( x(t) \) under a zero-mean prior, as in Equation \eqref{eq:gp-model}. At each timestep the model produces a full predictive distribution \( p(\epsilon \vert x(t)) \) rather than a point estimate, with posterior mean \( \mu(t) \) and standard deviation \( \sigma_{GP}(t) \).

\begin{equation}
\begin{aligned}
\label{eq:gp-model}
f(x(t)) \sim \mathcal{GP}(0, k(x(t), x'(t)))
\end{aligned}
\end{equation}

The chosen covariance is the Rational Quadratic (RQ) kernel, as in Equation \eqref{eq:rq-kernel}, where \( \sigma^2_f \) is the output scale, \( l_i \) is the ARD lengthscale of input dimension \( i \), and \( \alpha >0 \) is the scale-mixture parameter. This kernel suits the sensor response as it contains both fast local variations, such as hysteresis-loop crossings, and slower viscoelastic drift. ARD allows the model weight each of the three input dimensions according to its relevance for strain prediction.

\begin{equation}
\begin{aligned}
\label{eq:rq-kernel}
k(x, x') = \sigma^2_f \left(1 + \sum_{i=1}^{3} \frac{(x_i - x'_i)^2}{2 \alpha l^2_i}\right)^{-\alpha}
\end{aligned}
\end{equation}

\subsection{Risk Factor for Reliability Monitoring}\label{h2:reliability}

The GP posterior standard deviation \( \sigma_{GP}(t) \) is not enough to quantify the reliability of a sensor. This measurement only reflects the model's predictive variance and not whether the input is physically plausible or outside the training distribution. In this work, reliability is quantified as a risk factor \( p_{risk} \in \left[ 0,1 \right]  \) that combines two components: the probability of violating the physical constraints, and the degree to which the current input lies outside the training distribution. The risk factor is computed every timestep using the predicted mean \( \mu(t) \) and the standard deviation \( \sigma_{GP}(t) \). 

The first component covers, for each physical constraint \( k \) of Equation \eqref{eq:physical-constraint}, the probability that the constrained quantity \( q_k \), either the strain \( \epsilon \) or the strain rate \( \dot{\epsilon} \), exceeds its limit \( q_{\max,k} = \theta_k \). The quantity is treated as Gaussian with mean \( \mu_k(t) \) and standard deviation \( \sigma_k(t) \), so \( p_k(t) \) is the tail probability beyond \( q_{\max,k} \), as in Equation \eqref{eq:physical-violation-probability}.

\begin{equation}
\begin{aligned}
\label{eq:physical-violation-probability}
p_{k}(t) = P \left( \left\vert q_{k} \right\vert >q_{\max,k} \mid \mu_k(t), \sigma_k(t) \right)
\end{aligned}
\end{equation}

For the strain limit, \( \mu_k(t) = \mu(t) \) and \( \sigma_k(t) = \sigma_{GP}(t) \) are taken directly from the GP posterior. For the strain-rate limit, the strain rate and its uncertainty are obtained by differentiating the predicted strain and propagating the GP uncertainty at the two samples, as in Equation \eqref{eq:strain-rate-moments}, so that \( \mu_k = \dot{\mu} \) and \( \sigma_k = \dot{\sigma} \). This value grows when the model extrapolates beyond the training range.

\begin{equation}
\begin{aligned}
\label{eq:strain-rate-moments}
\dot{\mu}(t) = \frac{\mu(t) - \mu(t - \Delta t)}{\Delta t}, \quad \dot{\sigma}(t) = \frac{\sqrt{\sigma_{GP}^2(t) + \sigma_{GP}^2(t - \Delta t)}}{\Delta t}
\end{aligned}
\end{equation}

The second component is the epistemic risk. It measures how far the current predictive uncertainty \( \sigma_{GP}(t) \) is from the uncertainty observed over the nominal training data, as in Equation \eqref{eq:epistemic-risk}. When the input moves away from the training distribution, \( \sigma_{GP}(t) \) increases, and so does \( p_u \). This quantity indicates that the model is operating outside the region it was trained on and its predictions can no longer be trusted. In the equation, \( \sigma_{low} \) and \( \sigma_{high} \) are lower and upper bounds of the nominal training uncertainty distribution. In this work, \( \sigma_{low} \) is selected to be the median of the nominal training uncertainty (\( p_u=0 \)), and \( \sigma_{high} \) is set to the 99th percentile of the nominal training uncertainty (\( p_u=1 \) at the boundary).

\begin{equation}
\begin{aligned}
\label{eq:epistemic-risk}
p_u(t) = \operatorname{clip} \left( \frac{\sigma_{GP}(t) - \sigma_{low}}{\sigma_{high} - \sigma_{low}}, 0, 1 \right)
\end{aligned}
\end{equation}

The \( K \) physical components and the epistemic component are combined using the probability union formula to calculate the risk factor \( p_{risk} \), following Equation \eqref{eq:risk-factor}. Each component is independent, so any single one can drive \( p_{risk} \) towards its maximum value.

\begin{equation}
\begin{aligned}
\label{eq:risk-factor}
p_{risk}(t) = 1 - \left( 1 - p_{u}(t) \right) \prod_{k=1}^{K} \left( 1 - p_{k}(t) \right)
\end{aligned}
\end{equation}

\begin{figure*}[h!]
\centering
\begin{subfigure}[b]{0.49\linewidth}
\centering
\includegraphics[width=\linewidth]{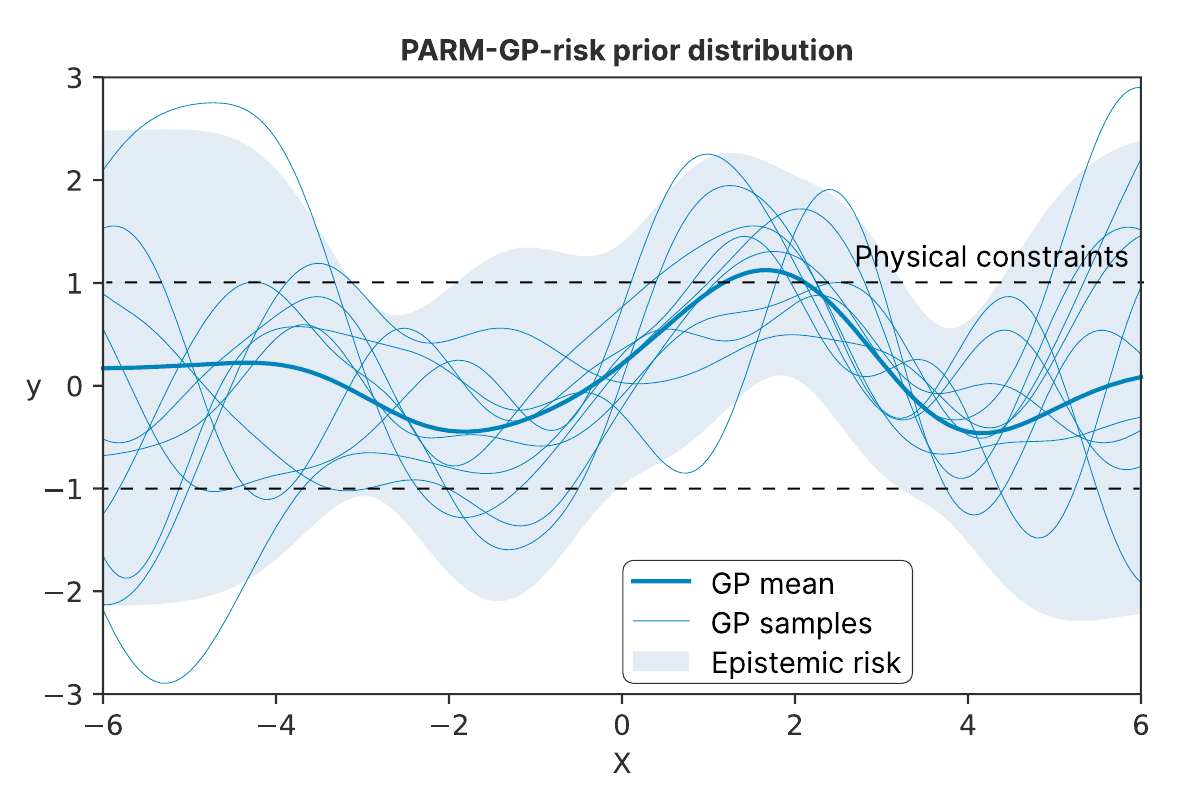}
\caption{}
\label{fig:parm-gp-prior}
\end{subfigure}%
\begin{subfigure}[b]{0.49\linewidth}
\centering
\includegraphics[width=\linewidth]{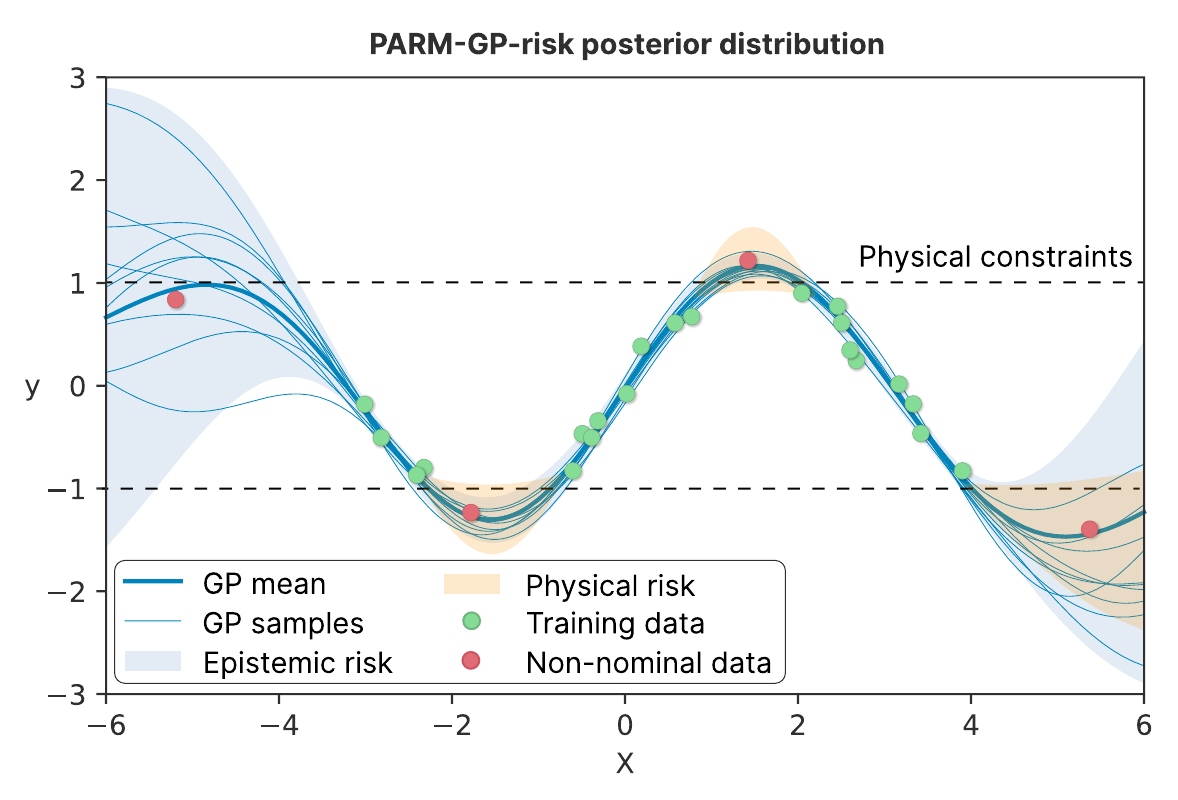}
\caption{}
\label{fig:parm-gp-posterior}
\end{subfigure}%
\\
\caption{Reliability monitoring methodology using the risk factor. (a) The GP network's uncertainty is high before training. (b) After training, GP uncertainty estimation within the training distribution x(t) decreases. For non-nominal data, risk increases either by exceeding physical constraints or by being outside the training distribution.}
\label{fig:pif-gp-risk}
\end{figure*}

Figure \ref{fig:pif-gp-risk} shows a graphical representation of the proposed physics-aware reliability monitor (PARM) and the the combination of the \( p_u \) and \( p_k \) probabilities. Before training, the GP prior exhibits high uncertainty and provides no information about the sensor's admissible physical domain. Training reduces the predictive uncertainty for inputs close to the training distribution. Nevertheless, some predictions remain highly confident while lying outside the admissible constraint bounds. Their risk therefore arises from violating the physical constraints rather than from epistemic uncertainty alone.

\section{Piezoresistive Soft Sensors Considered}

The proposed PARM framework is validated through two distinct soft piezoresistive sensor technologies. The superelastic Nitinol wire is proposed as the primary case study to analyze the method performance. Additionally, a silver-coated polyamide thread is included as a secondary validation example in the supplementary files to demonstrate that the proposed method is not tied to the specific physics of a determined sensor.

\subsection{Superelastic Nitinol Wire}

\begin{figure*}[htbp!]
\centering
\begin{subfigure}[b]{0.31\linewidth}
\centering
\includegraphics[width=\linewidth]{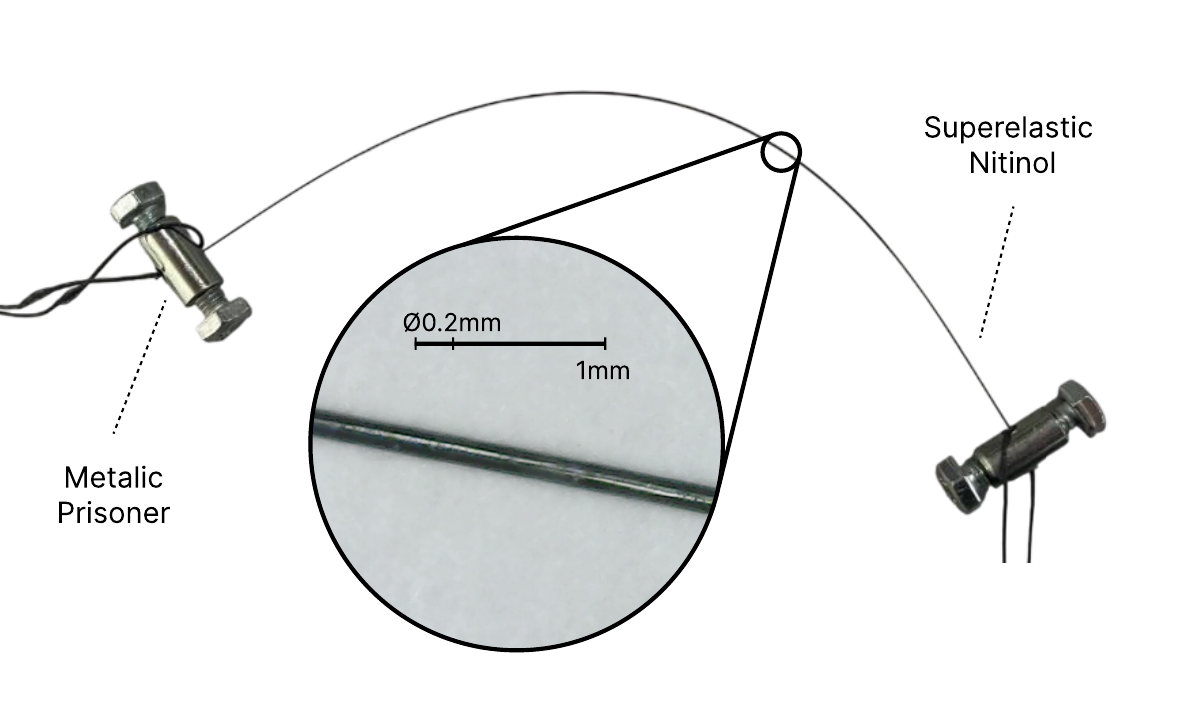}
\caption{}
\label{fig:se-picture}
\end{subfigure}%
\begin{subfigure}[b]{0.31\linewidth}
\centering
\includegraphics[width=\linewidth]{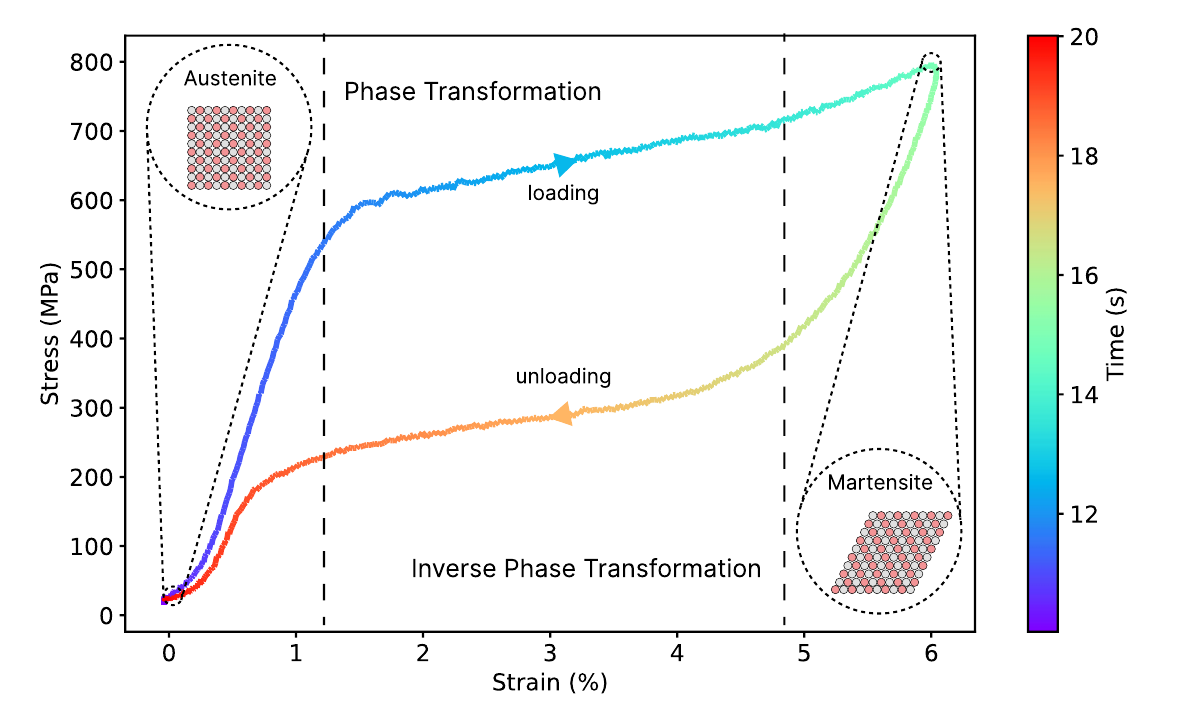}
\caption{}
\label{fig:se-hysteresis}
\end{subfigure}%
\begin{subfigure}[b]{0.31\linewidth}
\centering
\includegraphics[width=\linewidth]{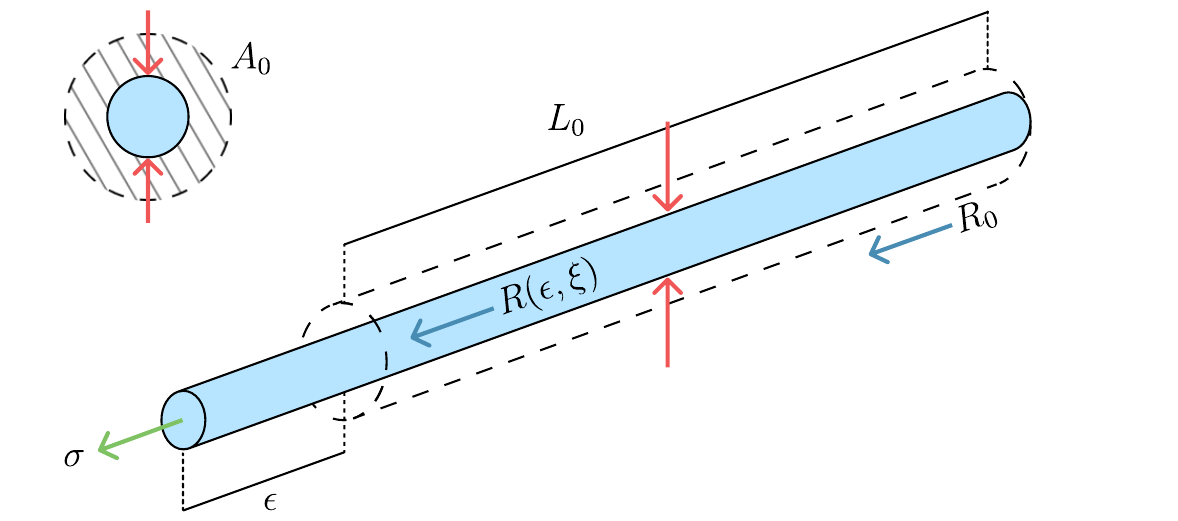}
\caption{}
\label{fig:se-poisson}
\end{subfigure}%
\\
\caption{Piezoresistive strain sensing material used for experimental validation. Superelastic Nitinol wire, including (a) the sensing element, (b) characteristic hysteresis behavior, and (c) geometric deformation.}
\label{fig:se-sensor}
\end{figure*}

The Nickel-Titanium alloy wire, often called Nitinol and shown in Figure \ref{fig:se-picture}, is a smart material that can be considerably stretched without producing any plastic deformation, unlike other metals. This property results from the transformation between two different atomic configurations, called phases, which is produced by either a change in temperature or applied external stress. When the temperature exceeds a certain value \( A_f \), the material is in its \textit{austenite} phase, the more symmetric atomic configuration. The other phase is \textit{martensite} and it is produced either when the temperature is below \( M_f \), or when an external stress is applied to the material while the temperature is above \( A_f \). The latter case produces stress-induced martensite, resulting in the superelasticity (SE) effect, where the material can undergo large strains and recover its original shape after unloading.

Figure \ref{fig:se-hysteresis} depicts the stress-strain diagram for the Nitinol alloy at a temperature above \( A_f \). Initially, the entire material is in its austenite phase, and when an external force is applied, the austenite deforms elastically with a resisitivity of \( \rho_A \) as the stress increases. At a certain point, the austenite begins to transform into martensite. During this transformation, the material is stretched and the strain increases, but the stress does not increase, producing the superelastic plateau. Once all the austenite has transformed into martensite, the martensite undergoes elastic deformation with a resistivity of \( \rho_M \) due to the change in its atomic structure. When the external force is removed, the process is inverted and the material returns to its original position. Since the reversible phase transformation follows two different energy paths during loading and unloading, hysteresis occurs in the material during stretching and release. 

Stretching the Nitinol wire produces a change in its resistance \( R (\epsilon,\xi) \) according to the Equation \eqref{eq:res-change}. The geometric part of the equation is produced by the Poisson Effect, shown in Figure \ref{fig:se-poisson}. When a wire is stretched in one direction, it elongates in the axially and contracts transversely. This term depends on the initial wire length \( L_0 \) and section \( A_0 \), and the strain \( \epsilon \) produced by the stretching. The piezoresistive part of the equation corresponds to the resistivity \( \rho(\xi) \), which also depends on the phase transformation of the material, because austenite and martensite have different microscopic configurations.

\begin{equation}
\begin{aligned}
\label{eq:res-change}
R(\epsilon, \xi) = \rho(\xi) \frac{L_0}{A_0}\left( 1+\epsilon \right)^2
\end{aligned}
\end{equation}

The material's amount in martensite phase is represented by the martensitic volume fraction \( \xi \in \left( 0,1 \right)  \). This value is used in Equation \eqref{eq:resistivity} to represent the percentage of phase transformation, obtaining the Nitinol's resistivity as a combination of the austenite resistivity \( \rho_A \) and the martensite resistivity \( \rho_M \).

\begin{equation}
\begin{aligned}
\label{eq:resistivity}
\rho(\xi) = \left( 1-\xi \right) \rho_A + \xi \rho_M
\end{aligned}
\end{equation}

The sensor's resistance is normalized to account for the relative changes with respect to an initial value \( R_0 \). The normalized value \( \tilde{R} \), defined by Equation \eqref{eq:res}, allows reproducibility regardless of initial resistance values that change depending on sensor length, integration or connectors. 

\begin{equation}
\begin{aligned}
\label{eq:res}
\tilde{R}(\epsilon,\xi) = \frac{R(\epsilon,\xi) - R_0}{R_0}
\end{aligned}
\end{equation}

\subsection{Data Acquisition}\label{h2:data-acquisition}

Table \ref{tab:conditions} shows the conditions \( C_{ij} \) of the reference signals for the static and the dynamic behavior of the given sensors. For each test condition, the sensor is calibrated at 1N beforehand. Then, the sensor is subjected to either 20 cycles of sinusoidal stretching to analyze the effect of drift and hysteresis across multiple cycles, or a cycle of step stretching to analyze the effect of fatigue over time. 

\begin{table}[htbp!]
\caption{Experimental dataset used for model training and validation. Static and dynamic loading conditions defined by strain amplitude and excitation frequency.}
\label{tab:conditions}
\begin{tabularx}{\linewidth}{lLL}
\toprule
\textbf{} & \textbf{Dynamic} & \textbf{Static} \\
\midrule
Signal type      & Sinusoidal                                    & Step \\
Strain amplitude & \( A \in \{1,...,6 \} \)\%                & \( A \in \{1,...,6 \} \)\% \\
Frequency        & \( f \in \{0.1k \vert k=1,...,10\} \)Hz & \( f = 0 \)Hz \\
Duration         & \( T_{tot} = \frac{20}{f} s \)                 & \( T_{tot} = 120 s \) \\
\bottomrule
\end{tabularx}
\end{table}

The dataset of the sensor's physical domain is composed by 66 files, each file containing time-series recordings of one experimental condition \( C_{ij} \). The time-series signals include the relative resistance \( \tilde R(t) \), the relative resistance rate \( \dot{\tilde R}(t) \), the Standard Linear Solid model memory state \( z(t) \), and the ground-truth strain \( \epsilon \), all sampled at a 100 Hz rate.

Three additional tests were recorded to evaluate sensor behavior when anomalies occur during nominal use: manually induced noise on the relative resistance, wire breakage, and operation with a fatigued wire.

\subsection{Sensor Performance}

The materials must be validated as suitable piezoresistive strain sensors before modeling them.

The Gauge Factor (\( GF \)) is a measurement of a sensor's relative sensitivity to strain. It determines how effectively a sensor converts physical deformation into a measurable electrical signal. It is defined as the ratio of relative resistance change to strain, according to the Equation \eqref{eq:gf}. A higher GF indicates that the sensor is better suited to detecting small, precise strains. 

\begin{equation}
\begin{aligned}
\label{eq:gf}
GF = \frac{\tilde{R}(\epsilon,\xi)}{\epsilon}
\end{aligned}
\end{equation}

The Pearson's correlation coefficient (\( r \)) is used to determine whether strain and resistance changes are related, as it measures the strength and direction of a linear relationship between two continuous variables. It is calculated as the ratio of the covariance of the variables to the product of their standard deviations, following Equation \eqref{eq:r}, where \( \bar{\epsilon} \) is the mean strain, \( \bar{\tilde{R}} \) is the mean relative resistance and \( N \) is the number of data samples. The result is a value ranging from \( r_{\epsilon, \tilde{R}} \in (-1, 1) \) where 0 represents no correlation and values close to -1 and 1 represent inverse and direct correlation, respectively. 

\begin{equation}
\begin{aligned}
\label{eq:r}
r_{\epsilon, \tilde{R}} = \frac{\sum_{i=1}^N (\epsilon_i - \bar{\epsilon})(\tilde{R}_i - \bar{\tilde{R}})}{\sqrt{\sum_{i=1}^N (\epsilon_i - \bar{\epsilon})^2} \sqrt{\sum_{i=1}^N (\tilde{R}_i - \bar{\tilde{R}})^2}}
\end{aligned}
\end{equation}

The Gauge Factor and Pearson correlation coefficient are evaluated across the dataset's 66 operating conditions. The Gauge Factor is computed empirically as the slope of a least-squares linear regression of \( \tilde{R} \) against \( \epsilon \) over the loading and unloading cycles, giving \( GF=5.98\pm 0.89 \), which indicates high sensitivity. Similarly, the Pearson correlation coefficient is \( r_{\epsilon,\tilde{R}} = 0.96\pm 0.04 \), showing a strong linear relationship between mechanical strain and resistance change, which supports accurate modelling of the sensor.

\section{Results}

The proposed PARM method is evaluated primarily on the superelastic Nitinol wire and then validated on the silver-coated polyamide thread, whose results appear in the supplementary file. The results cover the definition and analysis of the sensor's nominal region for optimized training, the performance of the trained model, and the reliability handling of the strain prediction using the risk factor assessment. 

\subsection{Nominal Region Identification}

The nominal region of the sensor is identified by imposing physical and quality constraints. The selected hard limits on strain and strain rate for the Nitinol wire are \( g_1(\epsilon(t)) = \epsilon(t) \leq 6\% \) and \( g_2(\dot{\epsilon}(t)) = \vert \dot{\epsilon}(t)\vert \leq 7 \%/s \). The quality constraint is based on the fit score defined in Equation \eqref{eq:fit-score}, using the threshold \( G_1 (C_{ij}) \geq 0.88 \), where \( \epsilon_k \) and \( \mu_k \) are the ground-truth and predicted strain over the \( N \) samples of condition \( C_{ij} \). This metric evaluates both dynamic and static signals on the same 0 to 1 scale, independently of the signal amplitude.

\begin{equation}
\begin{aligned}
\label{eq:fit-score}
G_1 \left( C_{ij} \right) = 1 - \sqrt{\frac{\sum_{k=1}^{N}(\epsilon_k - \mu_k)^2}{\sum_{k=1}^{N} \epsilon_k^2}}
\end{aligned}
\end{equation}

\begin{figure*}[htbp!]
\centering
\begin{subfigure}[b]{0.33\linewidth}
\centering
\includegraphics[width=\linewidth]{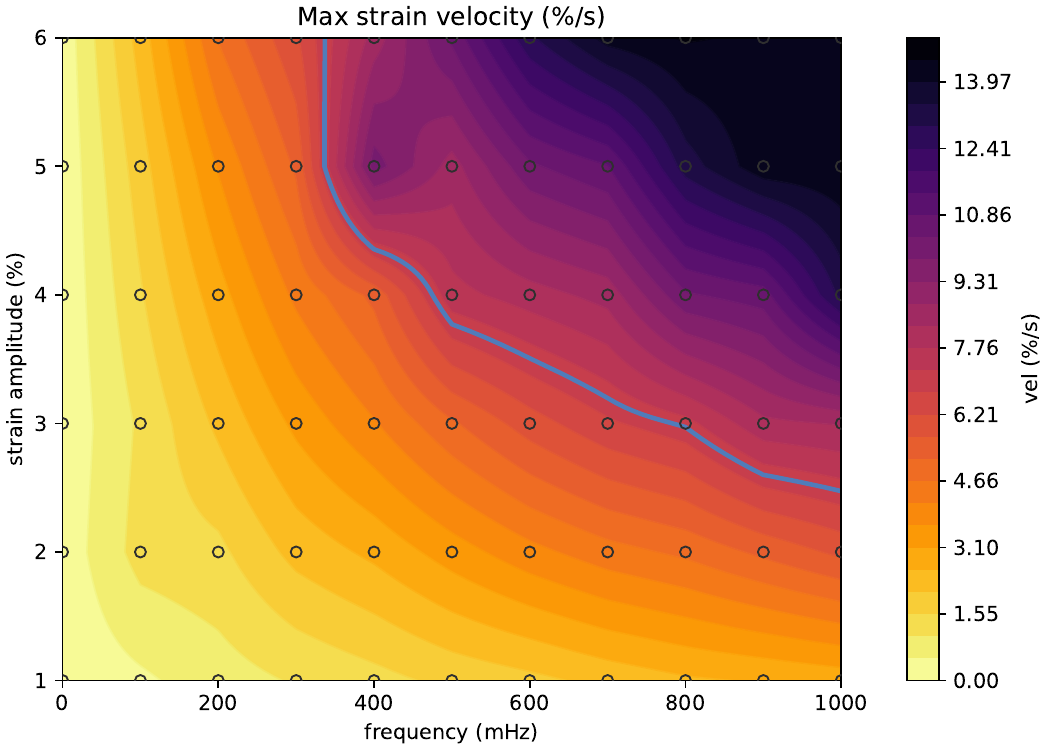}
\caption{}
\label{fig:workspace-vel}
\end{subfigure}%
\begin{subfigure}[b]{0.33\linewidth}
\centering
\includegraphics[width=\linewidth]{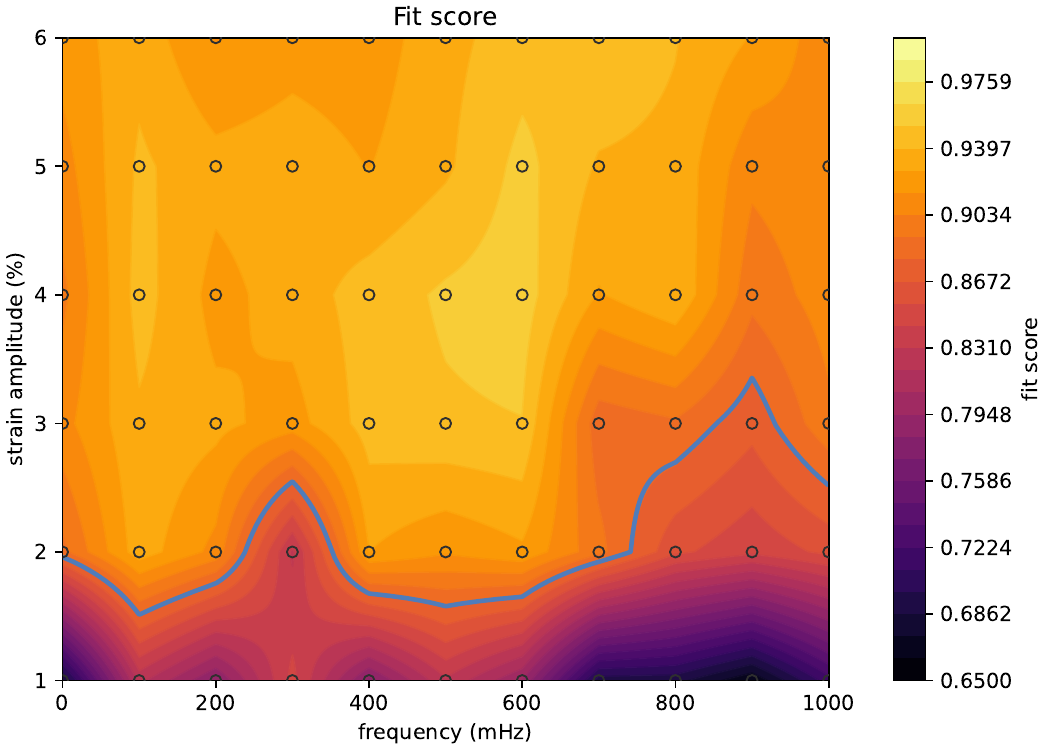}
\caption{}
\label{fig:workspace-r2}
\end{subfigure}%
\begin{subfigure}[b]{0.33\linewidth}
\centering
\includegraphics[width=\linewidth]{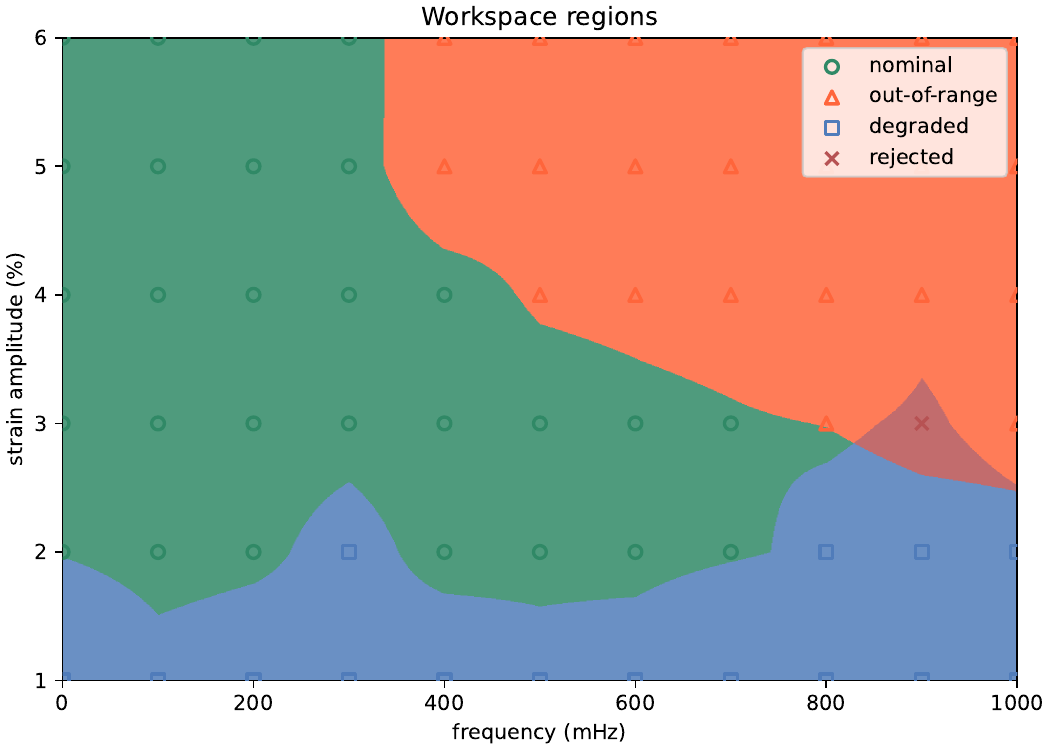}
\caption{}
\label{fig:workspace-mask}
\end{subfigure}%
\\
\caption{Definition of the operating workspace from the experimental dataset. (a) Maximum strain rate associated with each experimental condition. (b) Fit score across all loading conditions. (c) Resulting workspace partition into nominal, out-of-range, degraded, and rejected regions according to the selected physical constraints.}
\label{fig:workspace}
\end{figure*}

Figure \ref{fig:workspace} shows the results of applying the physical and quality constraints to the dataset of conditions for the Nitinol wire. Figure \ref{fig:workspace-vel} shows the heatmap of the maximum strain and strain rate and Figure \ref{fig:workspace-r2} shows the fit score for each condition. After applying the constraints, the physical domain of the sensor is divided into four regions, as shown in Figure \ref{fig:workspace-mask}. Low-strain conditions, particularly at 1\% amplitude, exhibit reduced fit scores because the resistance variation becomes comparable to measurement noise and hysteretic effects, making the inverse relationship less reliable. In contrast, high-frequency conditions exceed the imposed strain-rate limit.

\subsection{Uncertainty Calibration and Model Performance}

The risk factor's reliability depends on the model it runs on. A biased predictive mean shifts the physical-violation probability \( p_{k} \)'s tail estimate away from the true strain value, so a real out-of-range excursion can go undetected. A poorly calibrated model distorts the epistemic term \( p_{u} \) as well, since \( \sigma_{low} \) and \( \sigma_{high} \) are calibrated directly from the model's own nominal uncertainty distribution. This section selects the model's input feature set through an ablation over the candidate features \( x(t) \) and then checks the calibration of the resulting model.

Predictive accuracy is evaluated with Leave-One-Condition-Out Cross-Validation (LOCOCV) in the nominal region, training on all but one condition and testing on the condition left out, and again on out-of-range conditions using a model trained exclusively on nominal data. Both regions are scored with the fit score, shown in Equation \eqref{eq:fit-score}, and Root Mean Squared Error (RMSE) against the ground-truth strain. Calibration is scored separately, in the nominal region only, with the Prediction Interval Coverage Probability (PICP-95), as in Equation \eqref{eq:picp95}.

\begin{equation}
\begin{aligned}
\label{eq:picp95}
PICP_{95} = \frac{1}{N} \sum_{k=1}^{N} \mathds{1} \left(\vert \epsilon_k - \mu_k\vert \leq 1.96 \sigma_{GP,k} \right)
\end{aligned}
\end{equation}

Table \ref{tab:predictions-performance} reports two complementary evaluations, one for the nominal region and one for out-of-range conditions, across the four candidate feature sets from the ablation study. All values are the mean and standard deviation across the LOCOCV folds. Adding the relative resistance rate to the relative resistance alone raises the nominal fit score to 0.93 and lowers the RMSE to 0.16\%. The relative resistance rate captures part of the hysteresis that the instantaneous resistance misses on its own. The SLS memory state alone barely moves either metric. Combining it with the derivative gives the best nominal performance, a fit score of 0.95 and an RMSE of 0.15\%. The full feature set's PICP-95 also settles closer to the 95\% target, at 0.96 instead of 0.98. Its prediction intervals grow less conservative without losing coverage.

The same features carry over to out-of-range conditions, in the same order of importance. The relative resistance rate again gives the largest single gain, reaching a fit score of 0.86 and an RMSE of 0.46\%, down from 0.61\% for the relative resistance alone. The SLS state alone leaves the fit score close to baseline, though its RMSE comes out marginally higher, 0.63\% against 0.61\%. Combining all three features keeps the best fit score, 0.87, and matches the derivative's RMSE of 0.46\%. This out-of-range accuracy matters for reliability. A more accurate predictive mean reduces the likelihood of misleading strain estimates once the sensor leaves the nominal region. The risk factor separately flags these same measurements as lying outside the trusted range.

\begin{table*}[htbp!]
\caption{Predictive performance of the GP for different physics-guided input features. Mean and standard deviation of the fit score, RMSE, and PICP-95 obtained from Leave-One-Condition-Out Cross-Validation (LOCOCV) in the nominal region and evaluated on out-of-range conditions.}
\label{tab:predictions-performance}
\begin{tabularx}{\linewidth}{lLLLL}
\toprule
\textbf{Gaussian process} & \textbf{Region} & \textbf{fit score} & \textbf{RMSE (\%)} & \textbf{PICP-95} \\
\midrule
+ \( \tilde{R}(t) \)                                         & Nominal      & \(0.90\pm 0.06\) & \(0.26\pm 0.33\) & \(0.98\pm 0.02\) \\
& Out-of-range  & \(0.82\pm 0.07\) & \(0.61\pm 0.44\) & - \\
\cmidrule{2-5}
+ \( \tilde{R}(t) \), \( \dot{\tilde{R}}(t) \)              & Nominal      & \(0.93\pm 0.03\) & \(0.16\pm 0.08\) & \(0.98\pm 0.02\) \\
& Out-of-range  & \(0.86\pm 0.06\) & \(0.46\pm 0.26\) & - \\
\cmidrule{2-5}
+ \( \tilde{R}(t) \), \( z(t) \)                            & Nominal      & \(0.91\pm 0.10\) & \(0.27\pm 0.54\) & \(0.98\pm 0.02\) \\
& Out-of-range  & \(0.83\pm 0.12\) & \(0.63\pm 0.67\) & - \\
\cmidrule{2-5}
+ \( \tilde{R}(t) \), \( \dot{\tilde{R}}(t) \), \( z(t) \) & Nominal      & \(0.95\pm 0.03\) & \(0.15\pm 0.09\) & \(0.96\pm 0.03\) \\
& Out-of-range  & \(0.87\pm 0.07\) & \(0.46\pm 0.42\) & - \\
\bottomrule
\end{tabularx}
\end{table*}

\subsection{Reliability Performance}

The risk factor must handle two failure modes. The first one is detecting non-nominal conditions, as when the sensor's strain or strain rate exceeds its physical constraints. The second one is detecting anomalies that occur within the nominal region. This can happen when strain and strain rate stay inside their limits but the resistance signal no longer resembles anything seen during training. This section evaluates epistemic risk alone (\( p_u \)), physical risk alone (\( p_k \)), and their union (\( p_{risk} \), Equation \eqref{eq:risk-factor}) against both failure modes, using the same trained model in all three cases.

The risk factor is passed through a three-state machine with reliable, warning, and fault states, and it enters the warning state above 0.5 and the fault state above 0.75. A transition only takes effect after five consecutive samples in the new state, and an isolated spike in the raw risk signal does not by itself trigger a state change. State occupancy is then reported separately for nominal, out-of-range, and abnormal tests. Nominal tests check whether the risk factor stays in the reliable state while the sensor operates inside its nominal region. For out-of-range tests, the question is whether the risk factor detects conditions that violate the physical operating limits. Abnormal tests cover failures that can occur during otherwise valid operation, such as signal spikes, fatigue, or wire breakage. These tests are summarized with the Detection Rate (DR), as in Equation \eqref{eq:dr}, where \( N_{C} \) is the number of test files in the condition and \( Warning_i \) and \( Fault_i \) are the warning and fault state occupancy fractions for file \( i \), respectively.

\begin{equation}
\begin{aligned}
\label{eq:dr}
DR = \frac{1}{N_{C}} \sum_{i=1}^{N_{C}} \mathds{1} \left(Warning_i >0 \vee Fault_i >0 \right)
\end{aligned}
\end{equation}

The performance metrics of the risk factor state machine are shown in Table \ref{tab:risk-performance}. Under nominal conditions, the GP model achieves a \( 100\% \) reliable state occupancy and a DR of zero with every risk definition. This zero-false-alarm result is a second, independent piece of evidence for the calibration already shown in Table \ref{tab:predictions-performance}, where the nominal PICP-95 sat close to its 95\% target.

Epistemic risk alone detects only \( 40\% \) of the out-of-range conditions, since the GP's predictive uncertainty does not consistently cross the warning threshold for trajectories that still land close to the training data. This is consistent with the model still reaching a \( 0.87 \) fit score on those same conditions (Table \ref{tab:predictions-performance}): when the mean prediction stays accurate, the GP remains confident and the epistemic term rarely rises. Physical risk detects a much higher \( 92\% \) of the same conditions and remains in the warning state for \( 25\% \) of the time, against \( 2\% \) for epistemic risk alone. Combining both detects \( 95\% \), the highest of the three, with \( 19\% \) of that time in the warning state and \( 24\% \) in the fault state. The combined monitor catches most departures from the nominal region, even when the GP mean prediction remains accurate.

In abnormal scenarios, all variants report failure states, resulting in a DR of \( 100\% \). The type of state, however, depends on the risk source. Epistemic risk mainly produces fault states, which is expected when the input signal moves away from the nominal distribution. Physical risk produces more warning time, since abnormal signals can violate strain or strain-rate limits without always pushing the GP uncertainty to the fault threshold. The combined risk maintains the full detection rate while preserving the fault response of the epistemic component. Therefore, epistemic components improve the detection of anomalies within the nominal region, whereas physical constraints provide greater control in out-of-range situations by enforcing task-defined hard limits independently of the model's probabilistic uncertainty.

\begin{table*}[htbp!]
\caption{Reliability monitoring performance of the proposed risk factor. State occupancy and Detection Rate (DR) for nominal, out-of-range, and abnormal conditions using epistemic risk, physical risk, and their combination.}
\label{tab:risk-performance}
\begin{tabularx}{\linewidth}{llLLLL}
\toprule
\textbf{Risk factor} & \textbf{Region} & \textbf{Reliable (\%)} & \textbf{Warning (\%)} & \textbf{Fault (\%)} & \textbf{DR (\%)} \\
\midrule
\( p_u \)             & Nominal      & \(100\pm 0\) & \(0\pm 0\)   & \(0\pm 0\)   & 0 \\
& Out-of-range  & \(89\pm 21\) & \(2\pm 3\)   & \(9\pm 19\)  & 40 \\
& Abnormal     & \(69\pm 21\) & \(0\pm 0\)   & \(31\pm 21\) & 100 \\
\cmidrule{2-6}
\( p_k \)             & Nominal      & \(100\pm 0\) & \(0\pm 0\)   & \(0\pm 0\)   & 0 \\
& Out-of-range  & \(60\pm 29\) & \(25\pm 16\) & \(15\pm 15\) & 92 \\
& Abnormal     & \(69\pm 21\) & \(30\pm 21\) & \(0\pm 0\)   & 100 \\
\cmidrule{2-6}
\( p_{risk} \)        & Nominal      & \(100\pm 0\) & \(0\pm 0\)   & \(0\pm 0\)   & 0 \\
& Out-of-range  & \(57\pm 30\) & \(19\pm 12\) & \(24\pm 27\) & 95 \\
& Abnormal     & \(67\pm 20\) & \(2\pm 0\)   & \(31\pm 20\) & 100 \\
\bottomrule
\end{tabularx}
\end{table*}

\begin{figure*}[htbp!]
\centering
\begin{subfigure}[b]{0.49\linewidth}
\centering
\includegraphics[width=\linewidth]{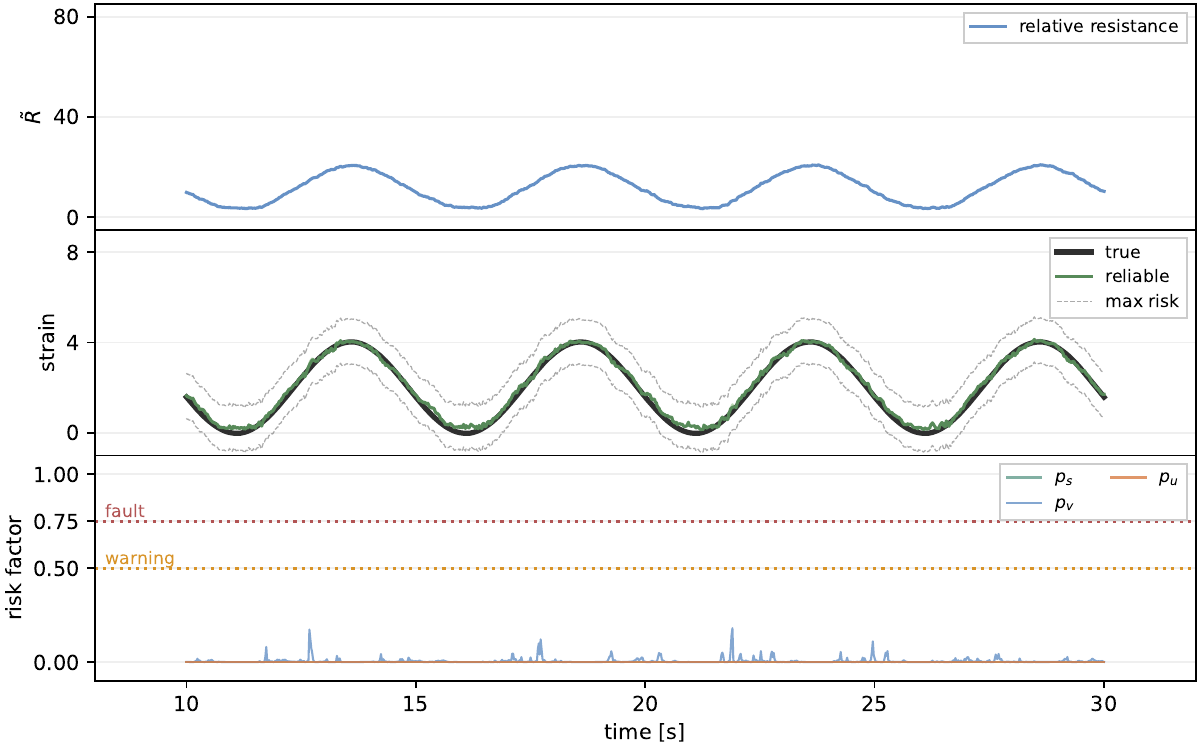}
\caption{}
\label{fig:nominal-sin-4-200}
\end{subfigure}%
\begin{subfigure}[b]{0.49\linewidth}
\centering
\includegraphics[width=\linewidth]{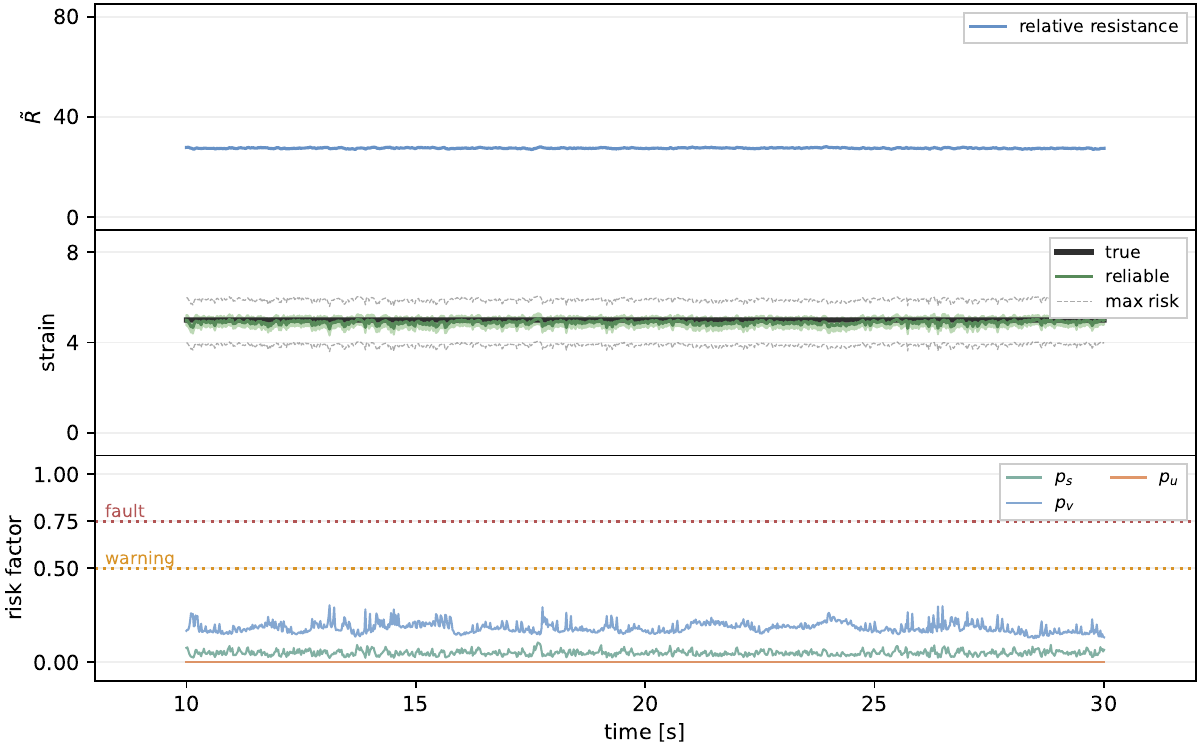}
\caption{}
\label{fig:nominal-step-6}
\end{subfigure}%
\\
\begin{subfigure}[b]{0.49\linewidth}
\centering
\includegraphics[width=\linewidth]{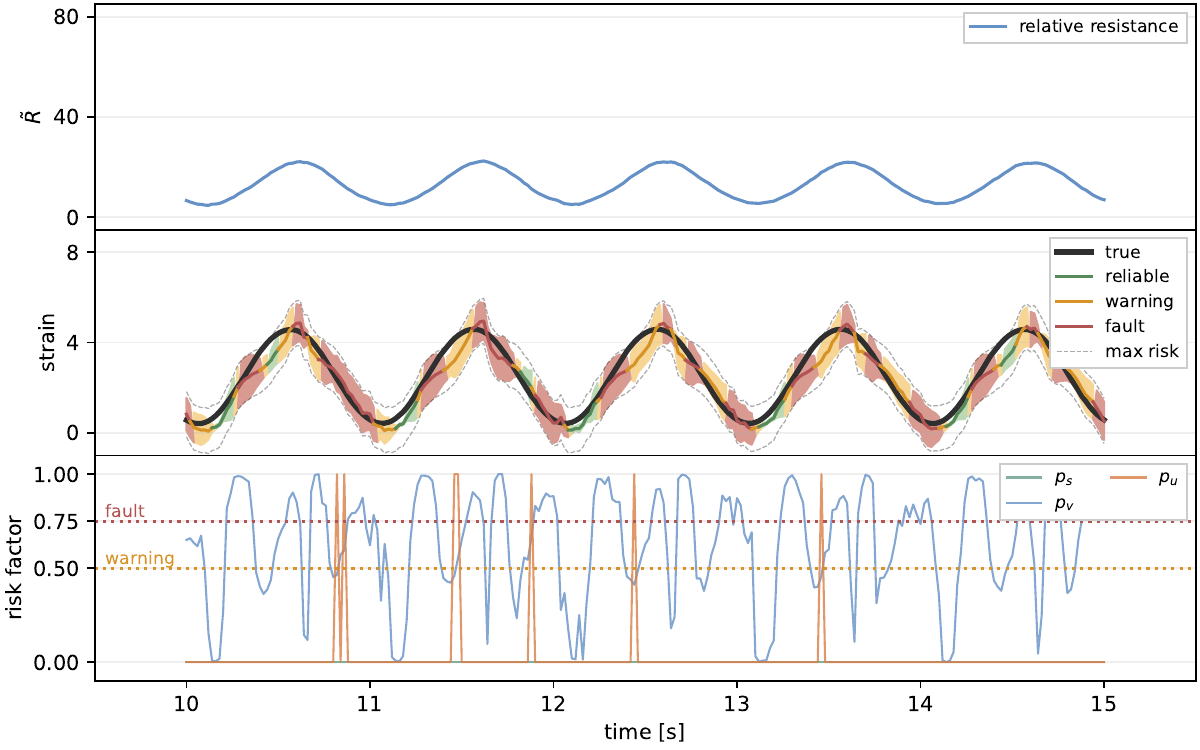}
\caption{}
\label{fig:oor-sin-5-1000}
\end{subfigure}%
\begin{subfigure}[b]{0.49\linewidth}
\centering
\includegraphics[width=\linewidth]{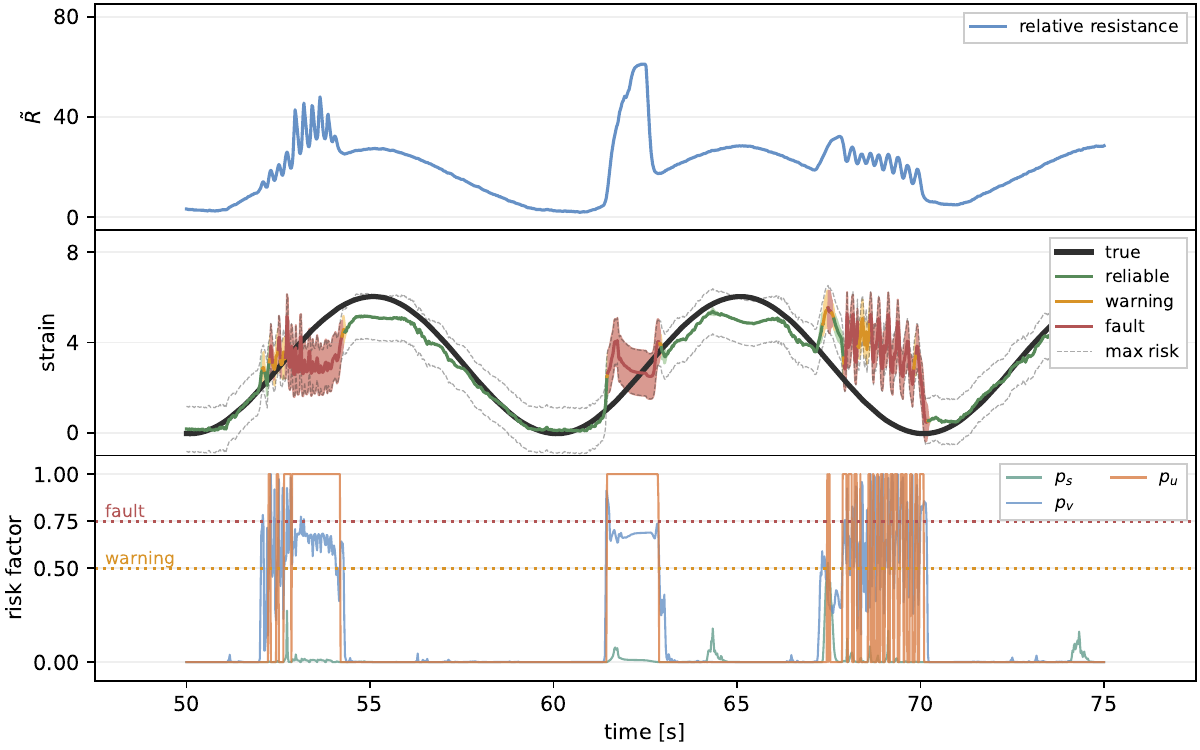}
\caption{}
\label{fig:abnormal-spikes-6}
\end{subfigure}%
\\
\begin{subfigure}[b]{0.49\linewidth}
\centering
\includegraphics[width=\linewidth]{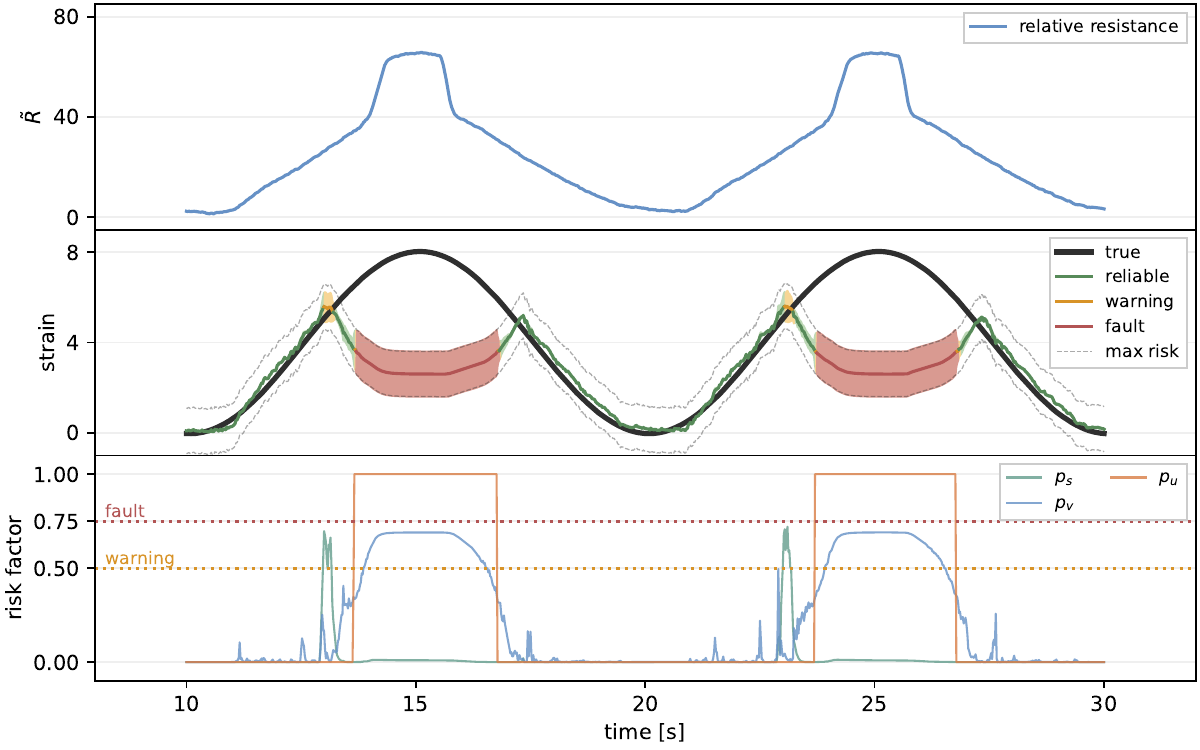}
\caption{}
\label{fig:abnormal-tetita-8}
\end{subfigure}%
\begin{subfigure}[b]{0.49\linewidth}
\centering
\includegraphics[width=\linewidth]{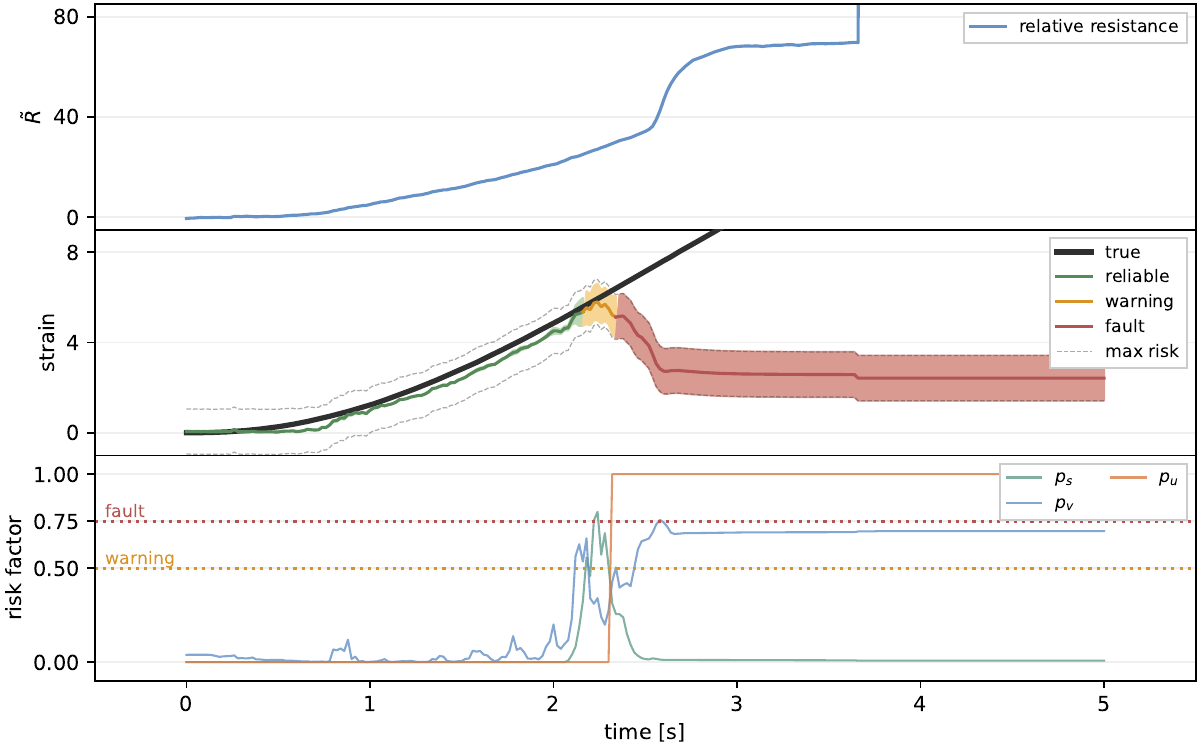}
\caption{}
\label{fig:abnormal-breakage}
\end{subfigure}%
\\
\caption{Representative strain estimation and reliability monitoring results. The measured relative resistance, probabilistic risk factor, reliability state, predicted strain, and ground-truth strain are shown for (a) nominal dynamic operation, (b) nominal static operation, (c) out-of-range operation, (d) abnormal signal spikes, (e) sensor fatigue, and (f) sensor breakage. The proposed framework maintains low risk within the calibrated operating region while detecting out-of-range operation and abnormal sensor behavior through the warning and fault states.}
\label{fig:predictions}
\end{figure*}

Figure \ref{fig:predictions} shows the time series of the relative resistance input, the GP strain prediction alongside the ground truth strain under different conditions, and the calculated risk factor together with the different probability components.

Figures \ref{fig:nominal-sin-4-200} and \ref{fig:nominal-step-6} depict the model's performance under nominal dynamic and static conditions, respectively. Accordingly, the risk factor remains very low because the input data lie within the GP training distribution, and the sensor state is confident in its prediction, which is consistent with the prediction accuracy and RMSE obtained.

Figure \ref{fig:oor-sin-5-1000} shows an out-of-range condition, in which the components of the risk factor operate together. When the input signals move far from the GP training data, the mean prediction relies on the model's generalization capability. Although the prediction remains reasonably accurate, its reliability decreases, resulting in an increased risk factor. The first component to be triggered is the strain-rate risk, indicating that the predicted strain rate has a high probability of exceeding the imposed strain-rate limit. Shortly afterward, the epistemic uncertainty of the GP also increases because the input data move away from the training distribution. Consequently, the out-of-range signal first triggers the warning state and then transitions to the fault state. This state machine enables the sensor to provide the system with continuous feedback about the reliability of its measurements, allowing failures that push the sensor beyond its nominal region to be detected.

Figure \ref{fig:abnormal-spikes-6} shows an erratic input signal that could be caused by an external disturbance. In this case, the risk factor rises rapidly, triggering the warning and fault states and providing feedback that disturbances are present in the sensor signal. Here, the main contributor to the risk alarm is the relative resistance rate, because the excitation condition remains within the nominal region and the predicted strain does not exceed the physical limit.

Another common failure mode of smart-material-based strain sensors is fatigue, which produces anomalous input signals that depend on the sensor's operating principle, as shown in Figure \ref{fig:abnormal-tetita-8}. In this case, cyclic loading modifies the phase transformation behaviour of the Nitinol wire through thermomechanical coupling, localized phase transformation, and accumulated microstructural damage. As a result, the relative resistance develops the characteristic bump near the maximum strain shown in the figure. The risk factor increases in response to this abnormal behaviour and identifies the sensor as faulty before mechanical failure or wire breakage occurs.

Finally, if the sensor breaks, which is also a common failure scenario, the circuit opens and a highly out-of-range input signal is produced, as shown in Figure \ref{fig:abnormal-breakage}. Traditional models would fail under these conditions, producing an unbounded strain estimate that could compromise the rest of the system. In contrast, the proposed PARM framework, together with the risk factor, places the sensor in the fault state, allowing the system to shut down and prevent further damage.

\section{Conclusions}

This paper proposed a reliability framework for piezoresistive soft sensors that pairs a physics-informed Gaussian Process inverse model with a risk factor combining predictive uncertainty and physical operating limits. Physics-guided input features raised the nominal fit score from 0.90 to 0.95, lowered RMSE to 0.15\%, and reached a 0.96 empirical coverage against the 95\% nominal target, and the same features reached a fit score of 0.87 when applied to out-of-range conditions. The Gaussian Process alone remained confident under those out-of-range conditions and caught only 40\% of them, so high predictive accuracy did not guarantee reliable detection. Fusing epistemic uncertainty with physical constraints raised detection to 95\% out-of-range and 100\% under abnormal operation, and nominal operation kept zero false alarms.

The resulting sensor reports a strain estimate together with a reliability state at every timestep, and this turns a conventional piezoresistive sensor into a self-monitoring one. The risk factor is computed from nominal operating data and physical limits alone, so no labeled fault examples are needed, and such examples are difficult to collect for soft sensing materials without deliberately damaging them.

The framework was validated on the Nitinol wire and on the silver-coated polyamide thread, two sensors with different transduction mechanisms, and the results show that the risk fusion rule is not tied to a specific material. The same rule extends to other physical constraints and other probabilistic regressors, since it only needs a predictive mean, a calibrated uncertainty, and a threshold on the quantity being bounded.

\bibliographystyle{IEEEtran}
\bibliography{refs}

\begin{IEEEbiographynophoto}{Carmen Ballester}
received her bachelor’s degree in Robotic Engineering from the University of Alicante in 2021, where she initiated her research career in 2019. She continued her academic pursuit, completing a Master’s in Robotics and Automation at Carlos III University of Madrid, where she became a member of the RoboticsLab in 2021. Currently a dedicated Ph.D. student in Electrical, Electronic, and Automation Engineering, Carmen specializes in wearable actuators and sensors within actuated exoskeletons for rehabilitation applications. Her expertise encompasses nonlinear control, soft robotics, smart materials, and rehabilitation robotics.
\end{IEEEbiographynophoto}

\begin{IEEEbiographynophoto}{Víctor Muñoz}
earned an Industrial Technology Engineering degree from the University of Málaga in 2021. He augmented his education with study programs in South Korea and completed his Master’s in Robotics and Automation at Carlos III University of Madrid in 2023. Starting his research career in 2021, he initially focused on mechanical ventilators at the University of Málaga. In 2022, he joined the RoboticsLab research group at Carlos III University, where he is currently a Ph.D. student specializing in Machine Learning applications in robotics.
\end{IEEEbiographynophoto}

\begin{IEEEbiographynophoto}{Dorin Copaci}
is an Assistant Professor in the Department of Systems Engineering and Automation at Carlos III University of Madrid, Spain. Since 2010, he has been a research member of RoboticsLab at the university. He completed his bachelor’s degree in Automatic Control and Systems Engineering at Politehnica University of Bucharest, Romania, in 2010. In 2012, he obtained his master’s degree in Robotics and Automation from Carlos III University in Spain, and he earned his Ph.D. degree in Electrical, Electronic, and Automatic Engineering in 2017. His research primarily focuses on the design and control of Shape Memory Alloy (SMA) actuators for rehabilitation devices.
\end{IEEEbiographynophoto}

\begin{IEEEbiographynophoto}{Dolores Blanco}
received the B.S. degree in physics from the University Complutense of Madrid, Spain, in 1992, and the Ph.D. degree in mechatronics from the University Carlos III of Madrid (UC3M), in 2002. In 1996, she joined the Department of Systems Engineering and Automation, UC3M, as a Fellowship Student, where she has been an Associate Professor since 2009. She is currently a member of the Robotics Laboratory Group, UC3M. Her current research interest includes emerging actuator technologies and design for rehabilitation robotics.
\end{IEEEbiographynophoto}

\end{document}